\documentclass[11pt]{article}
\usepackage[table]{xcolor}
\usepackage{colortbl}
\usepackage{arydshln}
\usepackage{multirow}
\usepackage{acl}
\usepackage{booktabs}
\usepackage{times}
\usepackage{latexsym}
\usepackage{comment}
\usepackage{booktabs}
\usepackage{amsmath,amssymb}
\usepackage{adjustbox}
\usepackage{graphicx}
\usepackage{float}

\usepackage{xcolor}

\usepackage{tcolorbox}
\tcbuselibrary{skins, breakable}

\usepackage{setspace}
\usepackage{enumitem}

\usepackage{amsmath}

\usepackage[T1]{fontenc}

\usepackage[utf8]{inputenc}

\usepackage{microtype}

\usepackage{inconsolata}

\usepackage{graphicx}

\title{Prompting Underestimates LLM Capability for Time Series Classification}

\author{Dan Schumacher, Erfan Nourbakhsh, Rocky Slavin, \and Anthony Rios \\
  The University of Texas at San Antonio \\
   \texttt{\{daniel.schumacher, anthony.rios\}}@utsa.edu \\}

\begin{document}
\maketitle
\begin{abstract}
Prompt-based evaluations suggest that large language models (LLMs) perform poorly on time-series classification, raising doubts about whether they encode meaningful temporal structure. We show that this conclusion reflects limitations of prompt-based generation rather than the model’s representational capacity by directly comparing prompt outputs with linear probes over the same internal representations. While zero-shot prompting performs near chance, linear probes improve average F1 from 0.15–0.26 to 0.61–0.67, often matching or exceeding specialized time-series models. Layer-wise analyses further show that class-discriminative time-series information emerges in early transformer layers and is amplified by visual and multimodal inputs. Together, these results demonstrate a systematic mismatch between what LLMs internally represent and what prompt-based evaluation reveals, leading current evaluations to underestimate their time-series understanding. 
\end{abstract}

\section{Introduction}

Time series analysis focuses on patterns in ordered sequences of events and is central to domains such as healthcare~\cite{AYALASOLARES2020103337, rajkomar_scalable_2018, choi2016doctor}, finance~\cite{fama1970efficient, Sirignano02092019}, and climate science~\cite{reichstein_deep_2019}. Time series classification plays a key role in these areas by assigning each sequence to a label based on its temporal structure and dynamics. Common applications include arrhythmia detection, human activity recognition, industrial fault detection, and environmental event identification.

LLMs demonstrate broad generalization across natural language tasks, supporting flexible pattern matching, contextual reasoning, and symbolic manipulation. These capabilities have motivated efforts to extend LLMs to non-linguistic data~\cite{fons-etal-2024-evaluating, Mai2024Opportunities}. Recent work has explored LLMs for time-series forecasting, classification, and analysis~\cite{jin2024timellmtimeseriesforecasting, gruver2024largelanguagemodelszeroshot, tan2024languagemodelsactuallyuseful, MoralesGarcia2025Developing}, often relying on major modifications and retraining. 
\emph{However, strong end-to-end performance in these settings does not necessarily imply that LLMs encode discriminative temporal structure, as results may instead reflect task formulation, additional fine-tuning, or prompting choices.}

\begin{figure}[t]
    \centering
    \includegraphics[width=.9\linewidth]{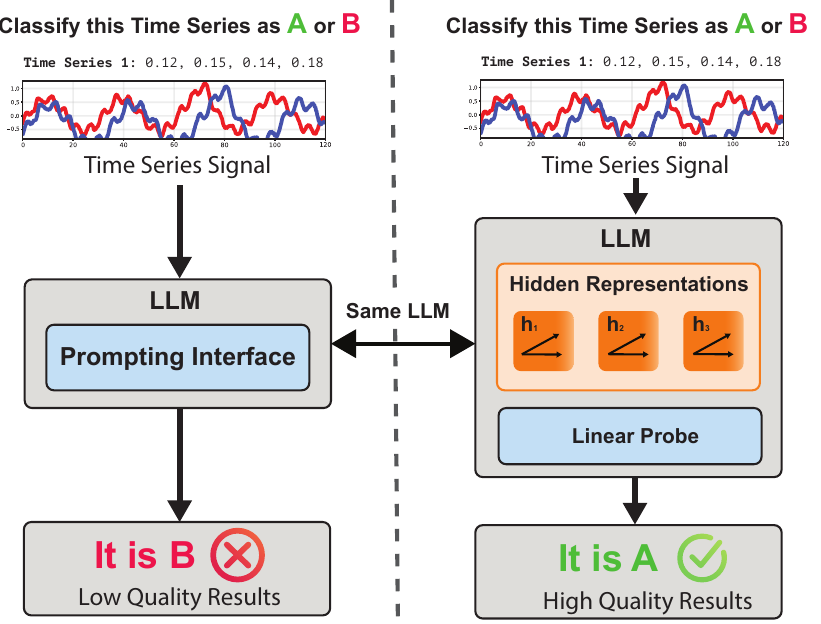}
    \caption{Does an LLM's \textit{observed} ability for time series classification match its underlying \textit{potential}?} \vspace{-1em}
    \label{fig:intro}
\end{figure}

\citet{merrill-etal-2024-language} show that, despite some success in zero-shot forecasting, LLMs struggle to reliably extract discriminative information from time series, performing near-randomly on reasoning and question-answering tasks that require interpreting temporal structure. Others similarly show that removing the LLM component from time-series pipelines often does not degrade performance and can even improve it, calling into question the necessity of language models for these tasks~\cite{tan2024languagemodelsactuallyuseful}. In contrast, a growing body of work argues that LLM-centric approaches may enable more general and interactive forms of time series understanding that extend beyond task-specific prediction~\cite{pmlr-v235-jin24i}. 
Together, these mixed findings raise a central question: \emph{do observed limitations reflect a lack of time-series understanding in LLMs, or a mismatch between what these models represent and how that information is extracted?}

Prior research that has attempted to use LLMs for time-series understanding has not directly assessed the intrinsic representational capacity of LLMs for time-series classification. Instead, performance is often conflated with three factors that obscure the underlying LLM's contributions. First, \emph{results can depend heavily on prompt engineering choices}, making it difficult to distinguish genuine temporal understanding from prompt-specific behavior. Second, \emph{many approaches rely on task-specific fine-tuning}, which alters the original pretrained model and prevents evaluation of the LLM's native capabilities. Third, LLMs are frequently combined with additional learned components such as auxiliary MLPs, transformer layers, or external embedding models, so \textit{improvements cannot be attributed to the language model itself}. As a result, existing evaluations typically treat LLMs as components within end-to-end systems, offering limited insight into whether out-of-the-box LLMs encode temporal patterns that are directly useful for classification.

In this paper, we evaluate prompting and hidden representation probing side by side to assess an LLM's \textit{observed} ability for time series classification versus its underlying \textit{potential} (see Figure~\ref{fig:intro}). We examine multiple input modalities to determine which are most effective for prompting and which yield representations that are most easily separable by a classifier. Compared with state-of-the-art methods as an upper bound, we find that simple linear probes are competitive with, and in some cases outperform, more complex LLM-based pipelines. Our results show that \emph{LLMs possess substantially greater time series classification representational capacity than is suggested by prompt-only evaluations}.  Moreover, \emph{we use linear probing as a diagnostic tool, restricting probe capacity and validating results against a random-weight model~\cite{chrupala-etal-2020-analyzing,tenney2019learncontextprobingsentence} baseline to rule out trivial memorization effects.}


To summarize, this work makes three primary contributions. \textbf{(1)} To the best of our knowledge, we are the first to explicitly disentangle prompt-level performance from representational capacity in time series classification. We do this by evaluating LLMs using both direct prompting and linear probing of their hidden states, which provides a clearer view of what temporal information is encoded versus what is accessible through generation. \textbf{(2)} We show that simple linear probes over hidden representations consistently match or outperform prompt-based and more complex LLM classification pipelines, indicating that much of the relevant time-series information is present but underused by prompting. \textbf{(3)} Through layer-wise analysis, we find that discriminative temporal structure emerges early in the model, suggesting that current limitations in LLM-based time series classification stem primarily from evaluation interfaces rather than a lack of temporal representations. Together, these results clarify when LLMs are best viewed as reasoning agents, feature extractors, or unnecessary components for time series classification.

\section{Related Work}

\vspace{2mm}
\noindent \textbf{Probing and Hidden Representations in LLMs.}
Early work on Layer-wise Relevance Propagation (LRP) introduced methods for attributing model predictions to intermediate representations~\cite{Bach2015Pixel} which later extended to more complex architectures~\cite{binder2016layerwiserelevancepropagationneural,arras-etal-2017-explaining}. Building on this work, probing methods emerged to explicitly measure what information these representations encode by testing their linear separability with lightweight classifiers~\cite{alain2018understandingintermediatelayersusing}.
In recent years, probing methods have increasingly been used as to extract features for a wide range of downstream applications. \citet{zhu-etal-2025-llm} and \citet{wang-etal-2025-faclens}, show that hidden states can be leveraged to estimate the difficulty an LLM is anticipated to have on a query prior to generation. More broadly, prior work has demonstrated that hidden representations encode rich and actionable signals that can be exploited across diverse settings, including alignment \citet{Kong2024Aligning, Li2023Inference, zhang2025realresponseembeddingbasedalignment}, safety \citet{wang2024probingsafetyresponseboundary, lu-etal-2025-x}, interpretability \citet{Ghandeharioun2024Patchscopes, jacobi2025superscopesamplifyinginternalfeature}, and robustness \citet{lad2024the, yan-etal-2024-contrastive}.

The most closely related work to ours is TiViT~\cite{roschmann2025timeseriesrepresentationsclassification}, which repurposes frozen vision and vision–language transformers for time-series tasks by first mapping time series into novel 2D image representations. This input transformation is central to the TiViT framework. In contrast, we study out-of-the-box LLMs and vision–language models (vLLMs), without relying on handcrafted time-series-to-image encodings. In addition, TiViT does not consider language models or prompt-based evaluation. Our work focuses on how different evaluation interfaces shape conclusions about model capability. We compare prompting and representation probing. For time series understanding, we study what representational power is gained through pretraining (relative to untrained models).

\vspace{2mm}
\noindent \textbf{Time-Series Research.}
Time series research spans a wide range of objectives, with forecasting and imputation forming the historical foundation of the field. Classical statistical methods, such as ARIMA and Kalman filters, model linear temporal dependencies~\cite{box1990TSA, hamilton1994time, hyndman2018forecasting}, while neural architectures, including LSTMs, convolutional networks, and transformers, enable richer nonlinear and long-range temporal modeling~\cite{hochreiter_long_1997, bai2018empiricalevaluationgenericconvolutional, Zhou_Zhang_Peng_Zhang_Li_Xiong_Zhang_2021, Wu2021Autoformer, LIM2021TemporalFusion}.

Building on this progression, recent work explores how large language models can be adapted for time series analysis. Existing approaches largely follow a small number of recurring patterns~\cite{Zhang2024Survey}: direct prompting that treats time series as text, often with modified numerical tokenization~\cite{gruver2024largelanguagemodelszeroshot, Hao2024PromptCast, liu2023largelanguagemodelsfewshot, xie2023wallstreetneophytezeroshot}; discretization or quantization schemes that map signals to token sequences~\cite{Chung2023TextToECG, Duan2023DeWave, Yang2024UniAudio, rubenstein2023audiopalmlargelanguagemodel, yu2023temporaldatameetsllm}; and architectures that introduce dedicated time-series encoders aligned with language model representations~\cite{King2023Multimodal, Sun2024test, zhou2023tentconnectlanguagemodels, qiu-etal-2023-transfer, Zhou2023OneFitsAll, Chang2025LLM4TS, jin2024timellmtimeseriesforecasting}. Related work also explores visual time-series representations processed by vision–language models or uses LLMs as controllers for external analysis tools~\cite{Girdhar2023ImageBind, su-etal-2023-pandagpt, moon-etal-2023-imu2clip, zhang2023insight}.

In contrast, we do not propose a new end-to-end system or optimize for task performance. Instead, we use probing as a diagnostic tool to isolate the representational capacity of pretrained LLMs and vLLMs for time series classification. By decoupling model representations from prompt engineering, fine-tuning, and auxiliary components, we provide a more direct analysis of what time-series structure these models encode.

\begin{figure*}[t]
    \centering
    \includegraphics[width=\textwidth]{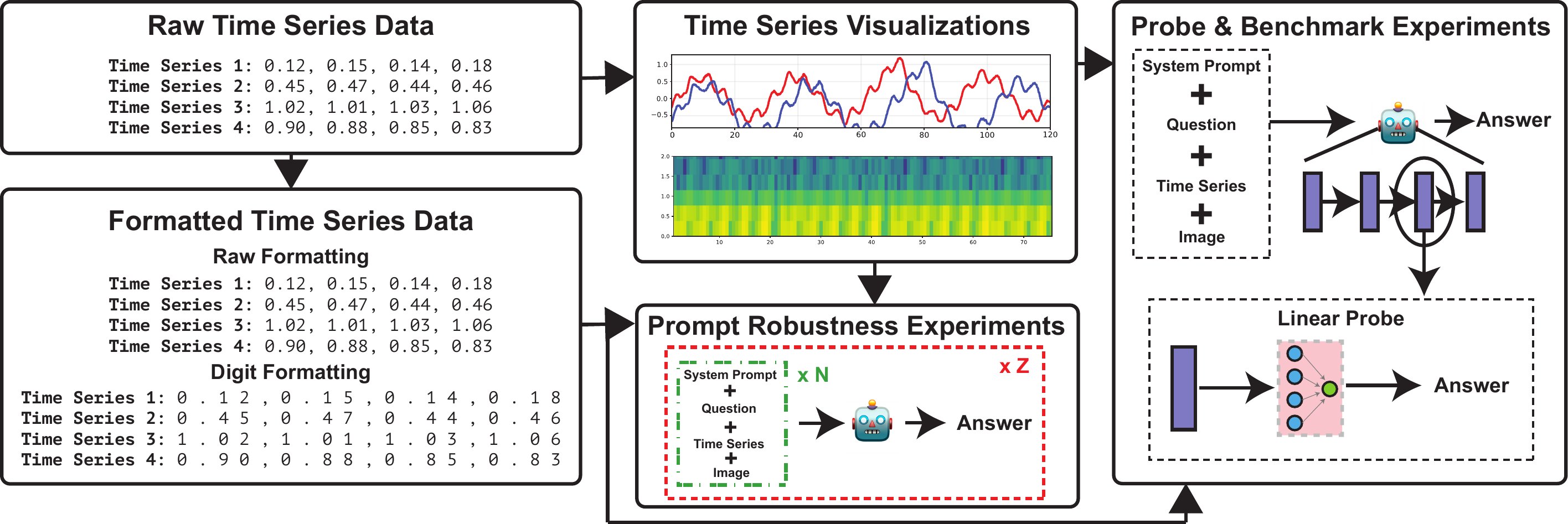}
    \caption{Overview of time series representations and prediction paradigms. Raw time series are transformed into digit-based text representations and visualizations. These representations are incorporated into prompts for direct prediction or used to extract hidden representations for layer-wise probing with linear classifiers.}\vspace{-1em}
    \label{fig:overview}
\end{figure*}

\section{Preliminaries and Problem Setup}

We study time series classification under LLM and vLLM interfaces. This section introduces the problem definition, the representations used to encode time series data, and the prediction paradigms evaluated in our study.

\subsection{Problem Definition}
Each time-series dataset is represented as a tensor $X \in \mathbb{R}^{N \times V \times T}$,
where $N$ denotes the number of sequences, $V$ the number of variables or channels,
and $T$ the temporal length of each sequence.
Time series classification is defined as follows:
given a training set $\{(X_i, y_i)\}_{i=1}^{N_{\text{train}}}$
and a test set $\{X_j\}_{j=1}^{N_{\text{test}}}$,
the objective is to predict class labels $y_j \in \{1, \ldots, C\}$
for each observation in the test set by identifying temporal patterns
that distinguish between classes.

\subsection{Time Series Representations}
We consider how LLMs and vLLMs perform with multiple representations of time series data. These representations are used consistently across both prompting and probing methods. The representations are described below:

\vspace{2mm} \noindent \textbf{Text Modality: Digit-Based Tokenization ($d$).}
Directly serializing numerical values can lead to unstable tokenization due to byte-pair encoding (BPE), which may fragment or merge digits unpredictably.
To mitigate this issue, we adopt a digit-based preprocessing strategy from \citet{gruver2024largelanguagemodelszeroshot}.
Each numerical value is converted into a digit-space format with fixed precision, where commas separate time steps and decimal points are removed.
For example, the time series \texttt{[1.0, 20, 0.33]} is transformed into
\texttt{"1 0 0 , 2 0 0 0 , 0 3 3"}.
This representation ensures that each digit maps to a consistent token and preserves numerical structure without BPE-induced fragmentation.

\vspace{2mm} \noindent \textbf{Visual Representation ($v$).}
Each time series is also rendered as an image, using either line plots or spectrograms, depending on the dataset.
Plots include appropriate titles, axis labels, and legends for multivariate data.
These visualizations provide an alternative modality that leverages the strengths of vLLMs for pattern recognition.
Additional details can be found in Appendix~\ref{sec:appendix:prompt_examples:ex_plots}.

\vspace{2mm} \noindent \textbf{Multimodal Representation ($d + v$).}
Finally, we evaluate a multimodal representation that combines digit-based text input with visual input, enabling simultaneous access to numerical and visual structure. 

\subsection{Prediction Paradigms}
We study two complementary approaches to time series classification: direct prompt-based prediction and representation probing. Figure~\ref{fig:overview} provides an overview of the representations and prediction pipelines evaluated in this work.

\vspace{2mm} \noindent \textbf{Prompt-Based Classification.}
For a given test time series $X_j$, we construct a prompt
$p_j = \tau + q + m$,
where $\tau$ denotes the task description,
$q$ is a multiple-choice question,
and $m \in \{d, v, d{+}v\}$ specifies the input modality.
A model $f$ (either an LLM or vLLM) then produces a predicted label
$y_j \in \{1, \ldots, C\}$ via text generation.
Formally,
$
y_j = f(p_j), \quad \text{where } p_j = \phi(X_j, \tau, q, m).
$
Please see the Appendix~\ref{sec:appendix:prompt_examples} for a listing of all prompts.

\vspace{2mm} \noindent \textbf{Representation Probing.}
Using the same prompt $p_j$, we additionally evaluate whether class-relevant information is encoded in the model’s internal representations.
Rather than using the generated output, we extract hidden states from the model.
For each layer $\ell \in \{1, \ldots, L\}$, we obtain an embedding
$h_j^{(\ell)} \in \mathbb{R}^{D}$, taken from the final-token hidden state.

For each layer, we train an independent logistic regression classifier using the corresponding embeddings.
Probes are trained with 5-fold cross validation and a maximum of 1000 iterations to select the regularization parameter $C$.
Formally,$h_j^{(\ell)} = \psi_\ell\!\big(f, p_j\big), \quad \ell \in \{1,\ldots,L\},$
and predictions are given by $\hat{y}_j^{(\ell)} = g_\ell\!\left(h_j^{(\ell)}\right),$
where $g_\ell$ denotes a linear probe trained on
$\{(h_i^{(\ell)}, y_i)\}_{i=1}^{N_{\text{train}}}$ and evaluated on
$\{h_j^{(\ell)}\}_{j=1}^{N_{\text{test}}}$.

Probing classifiers'  interpretation are sensitive to probe complexity, baselines, and the correlational nature of the analysis~\cite{belinkov_probing_2022}; we address these concerns with additional controls and analyses in Appendix~\ref{sec:appendix:probing}.

\section{Experiments}


We evaluate prompting and probing on multiple time-series classification datasets (described in Appendix~\ref{sec:appendix:BDMM:A3Datasets}, descriptive statistics found in Appendix~\ref{sec:appendix:stats}) and compare them against a range of heuristic, random, and state-of-the-art baselines (Described in Appendix~\ref{sec:appendix:baselines}).

\begin{table*}[t]
\centering
\footnotesize

\begin{tabular}{llrrrrrrr}
\hline
\toprule
\textbf{Model}& \textbf{Method}& \textbf{CTU}& \textbf{EMG}& \textbf{HAD}& \textbf{HAR}& \textbf{RWC}& \textbf{TEE}& \textbf{Avg}\\ \midrule
Attend       &  TS-Encoder          & .848 & .985 & .732 & .933 & .726 & .524 & .791       \\
TS2Vec&  TS-Encoder          & .631 & .933 & .667 & .865 & .672 & .794 & .761       \\
Moment       &  TS-Encoder          & .660 & 1.00& .671 & .878 & .778 & .564 & .758       \\
OneFitsAll   &  Encoder Alignment   & .587 & .296 & .681 & .885 & .790 & .475 & .615       \\
InstructTime &  TS-Quant + Prompt   & .627 & .167 & .278 & .527 & .434 & .235 & .378       \\
\midrule
 \multirow{4}{*}{Random}       & Probe  & .573 & .933 & .369 & .592 & .597 & .436 & .583       \\
             & Majority                 & .333 & .167 & .015 & .048 & .434 & .036 & .195       \\
             & Prior                    & .439 & .265 & .088 & .166 & .504 & .072 & .282       \\
             & Uniform& .442 & .333 & .084 & .164 & .463 & .141 & .296       \\
 & Digit Reg& .487& 0.463& 0.134& 0.488& 0.497& 0.399&.411\\
\midrule
 \multirow{2}{*}{Llama}        & Probe   & .640 & 1.00& .406 & .723 & .672 & .223 & \textbf{.611}           \\
             & Prompt                    & .356 & .175 & .033 & .111 & .190 & .036 & .150       \\ \cmidrule{2-9}
 \multirow{2}{*}{Mistral}        & Probe & .684& .933& .438& .730& .645& .462& \textbf{.649}\\
             & Prompt                    & .548 & .393 & .021 & .127 & .190 & .288 & .261       \\\cmidrule{2-9}
 \multirow{2}{*}{Qwen}           & Probe & .676 & 1.00& .374 & .737 & .656 & .592 & \textbf{.672}       \\
             & Prompt                    & .324 & .413 & .016 & .174 & .222 & .304 & .242            \\
             
\bottomrule
\end{tabular}
\caption{F1 scores across all datasets. Prompt and Probe results pertain only to the $d+v$ modality. Results are organized into three groups: (top) established time-series baselines; (middle) random and heuristic baselines; and (bottom) vLLM Probe and Prompting. Established baselines indicate strong upper-bound performance, while random methods provide a lower-bound reference. } \vspace{-1em}
\label{MainResults}
\end{table*}


\vspace{2mm} \noindent \textbf{Main Findings.}
Table \ref{MainResults} summarizes the performance of probing and prompting methods relative to state-of-the-art and heuristic baselines across all evaluation datasets. The Majority, Prior, and Uniform baselines achieve average F1 scores of .195, .282, and .296, respectively. These results define absolute lower bounds for random classification. The Random-Probe baseline (More details in Appendix~\ref{sec:appendix:probing}) achieves an average F1 of .583. This score represents a practical floor for probing-based approaches and reflects how well a linear classifier can extract signal from features without meaningful representations. All probe-based models exceed the Random-Probe baseline, with average F1 scores of .611, .649, and .672. This result alleviates concerns that performance gains are driven solely by supervised learning. Additionally, Table~\ref{ProbeModalities} shows that the performance gap between Random-Probe and vLLM-Probe is substantially larger in other modalities.

\emph{vLLMs encode useful time-series features but struggle to exploit them during prompting.}
Across all models, prompting performs substantially worse than probing. For Llama, average F1 increases from .150 with prompting to .611 with probing. Mistral improves from .261 to .649. Qwen improves from .242 to .672. This pattern holds consistently across every dataset and every model. For each case, the probing score exceeds the corresponding prompting score. This effect is most pronounced on the EMG dataset. EMG appears to be an easy classification benchmark. Attend achieves an F1 of .985, and Moment, Llama, and Qwen all achieve perfect separation when probed. Despite this, prompting severely underestimates model capability on this dataset. Llama improves from .175 with prompting to 1.00 with probing. Mistral improves from .393 to .933. Qwen improves from .413 to 1.00. These gaps indicate that the learned representations are strong, but current prompting strategies fail to reliably access them.

\begin{figure}[t]
    \centering
    \includegraphics[width=.9\linewidth]{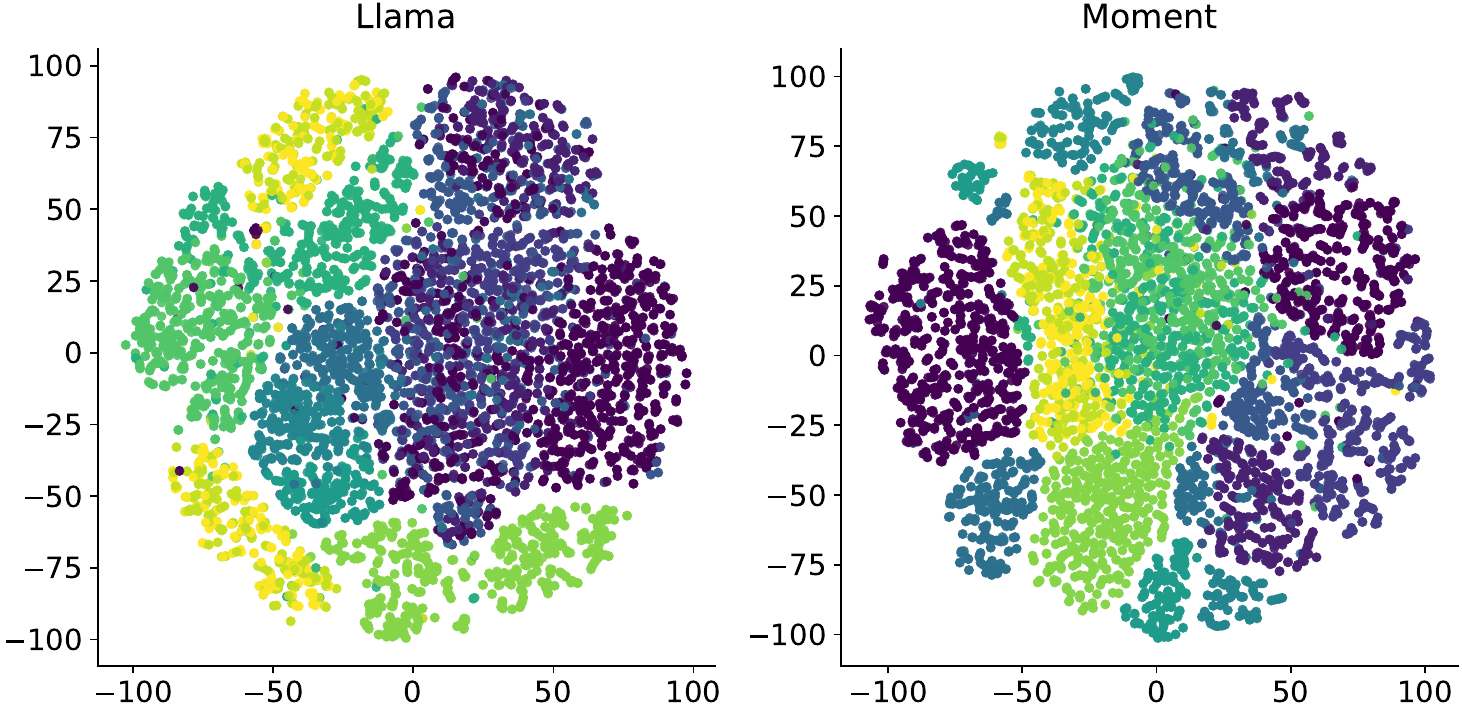}
    \caption{t-SNE visualizations of probe embeddings for Llama (left) and a time-series baseline (Moment, right) on the HAD dataset, illustrating comparable class separation.}\vspace{-1em}
    \label{fig:TSNE}
\end{figure}

\emph{vLLMs sometimes outperform state-of-the-art baselines.}
vLLM probing results are competitive with, and sometimes exceed, established time-series baselines. Qwen (.672) and Mistral (.649) outperform OneFitsAll (.615). Llama narrowly trails at .611. All three vLLMs outperform InstructTime. On individual datasets, vLLMs achieve state-of-the-art performance in several cases. On EMG, Llama and Qwen achieve perfect F1 scores of 1.00, exceeding Attend (.985) and TS2Vec (.933), and matching Moment. On CTU, Llama (.640) and Qwen (.676) outperform all state-of-the-art baselines except Attend. Figure~\ref{fig:TSNE} shows two t-SNE plots comparing Llama $v$ embeddings with Moment embeddings on the HAD dataset. This visualization highlights differences in feature separability and illustrates Llama's usefulness for classification compared to an established baseline.

\begin{table*}[t]
\centering
\footnotesize
\renewcommand{\arraystretch}{1.08}
\begin{tabular}{@{}llllllllll@{}}
\toprule
\textbf{Model} & \textbf{Method} & \textbf{Modality} &
\textbf{CTU} & \textbf{EMG} & \textbf{HAD} &
\textbf{HAR} & \textbf{RWC} & \textbf{TEE} & \textbf{Avg} \\
\midrule

\multirow{6}{*}{Llama}
 & \multirow{3}{*}{Probe}
 &          $d$   & .616 & .866 & .313 & .621 & .626 & .335 &   .547    \\
 &        & $v$   & .718 & .933 & .468 & .799 & .690 & .512 &   \textbf{.678}    \\
&        &  $d+v$ & .654 & 1.00 & .376 & .708 & .672 & .223 &   .597    \\

 \cmidrule{3-10}
 & \multirow{3}{*}{Prompt}
 &          $d$   & .333 & .171 & .032 & .054 & .104 & .036 & .132       \\
 &        & $v$   & .333 & .159 & .024 & .077 & .190 & .036 & .132       \\
  &       & $d+v$ & .344 & .171 & .033 & .111 & .190 & .036 & .160       \\

\midrule

\multirow{6}{*}{Mistral}
 & \multirow{3}{*}{Probe}
 &          $d$   & .682 & .712 & .386 & .735 & .628 & .265 &   .579    \\
 &        & $v$   & .696 & .796 & .352 & .684 & .664 & .439 &   .596    \\
  &        &$d+v$ & .698 & .933 & .393 & .729 & .645 & .462 &   \textbf{.631}    \\

 \cmidrule{3-10}
 & \multirow{3}{*}{Prompt}
 &          $d$   & .402 & .280 & .030 & .153 & .497 & .260 & .272       \\
 &        & $v$   & .445 & .663 & .055 & .225 &      .460& .302 &            .354\\
  &        &$d+v$ & .511 & .280 & .021 & .173 & .190 & .200 & .254       \\

\midrule

\multirow{6}{*}{Qwen}
 & \multirow{3}{*}{Probe}
 &          $d$   & .672 & .933 & .361 & .764 & .647 & .419 & .633       \\
 &        & $v$   & .688 & .866 & .304 & .655 & .640 & .575 & .621       \\
  &        &$d+v$ & .676 & 1.00 & .374 & .737 & .656 & .592 & \textbf{.672}       \\

 \cmidrule{3-10}
 & \multirow{3}{*}{Prompt}
 &          $d$   & .393 & .372 & .039 & .146 &      .432& .159 &            .259\\
 &        & $v$   & .341 & .618 & .028 & .220 & .296 & .246 & .326       \\
  &        &$d+v$ & .324 & .459 & .016 & .174 &      .222& .194 &            .255\\

\midrule

\multirow{3}{*}{Random}
 & \multirow{3}{*}{Probe}
 &          $d$   & .587 & .861 & .366 & .583 & .605 & .433 & .572       \\
 &        & $v$   & .333 & .167 & .015 & .048 & .434 & .036 & .172       \\
  &        &$d+v$ & .573 & .933 & .369 & .592 & .597 & .436 & .583       \\

\bottomrule
\end{tabular}
\caption{F1 for each dataset split by model, method, and modality. All probing results are \textit{much higher} than their corresponding prompt results.} \vspace{-1em}
\label{ProbeModalities}
\end{table*}

\emph{Probing and prompting benefit from visual information.}
Table~\ref{ProbeModalities} reports performance across input modalities for both probing and prompting. Probing performance generally improves with the inclusion of $v$. Llama performs best when using $v$, achieving an average F1 of .678. Mistral and Qwen perform best with $d+v$, achieving average F1 scores of .631 and .672, respectively. Prompting performance also improves with the inclusion of visual input, particularly for Mistral and Qwen. For Mistral, the $v$ modality achieves the highest average F1 at .354, followed by $d$ at .273. For Qwen, $v$ again performs best with an average F1 of .326, followed by $d$ at .259. This effect is most pronounced on the EMG dataset. For Mistral, prompting with $v$ achieves an F1 of .663, compared to .280 for the other modalities. For Qwen, prompting with $v$ achieves .618, compared to .372 with $d$ and .459 with $d+v$. This subset of results (Mistral \& Qwen + EMG + $v$) is the closest prompting comes to matching probing performance bound by the same modality ( Mistral $v$ at .796, Qwen $v$ at .866).

\emph{Random-Probe is a strong baseline.}
The Random-Probe baseline with $d+v$ achieves an average F1 of .583. This score exceeds the probing performance of Llama with $d$ (.547) and Mistral with $d$ (.579). This result indicates that a substantial portion of probing performance can be attributed to feature structure alone, independent of semantic alignment. It reinforces that probing gains are not solely an artifact of supervised learning. At the same time, it raises questions about how much semantic understanding is required to achieve competitive performance.

\emph{Text-only input is more robust than expected.}
Text-only input $d$ performs surprisingly well across models and methods. For Qwen, probing with $d$ (.633) is close to its best-performing modality (.672). For Mistral, probing with $d$ (.579) remains competitive with $v$ (.596). Even in prompting, $d$ frequently outperforms the combined $d+v$ modality. This suggests that the textual representation captures meaningful time-series structure.

\begin{figure}[t]
    \centering
    \includegraphics[width=\linewidth]{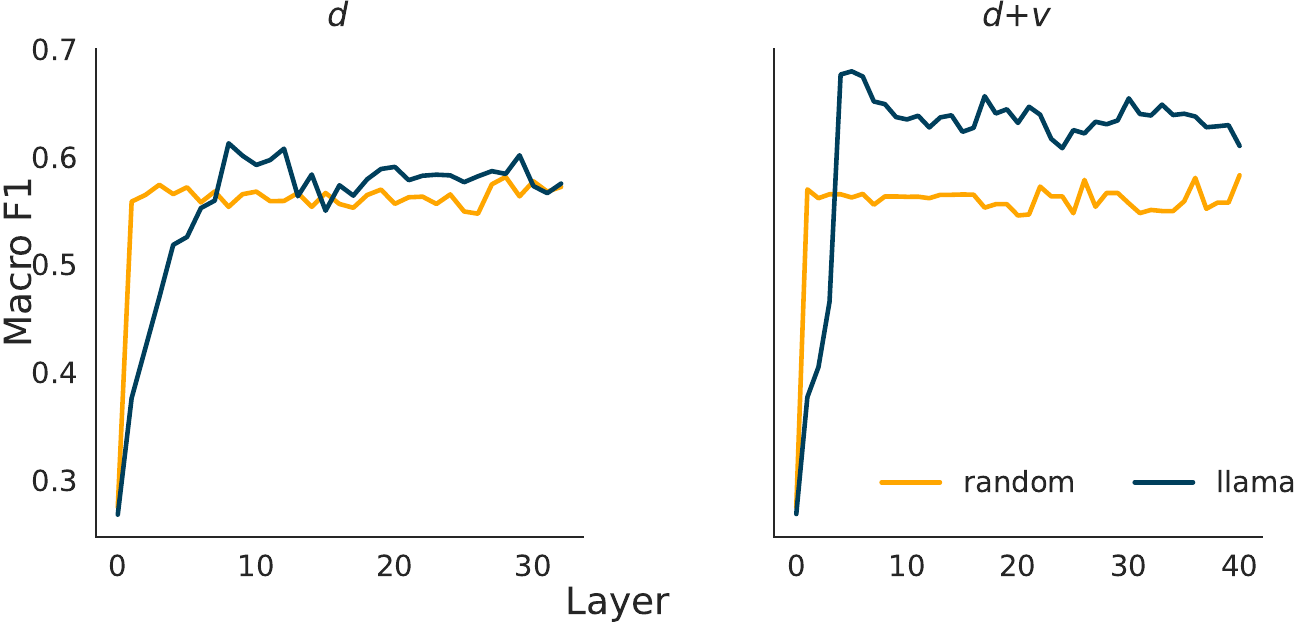}
    \caption{Layer-wise macro F1 of linear probes for Llama versus a random baseline under digit-only (d) and multimodal (d+v) inputs, showing early emergence of discriminative time-series information.}\vspace{-0em}
    \label{fig:RandomLlamaProbe}
\end{figure}

\begin{figure}[t]
    \centering
    \includegraphics[width=\linewidth]{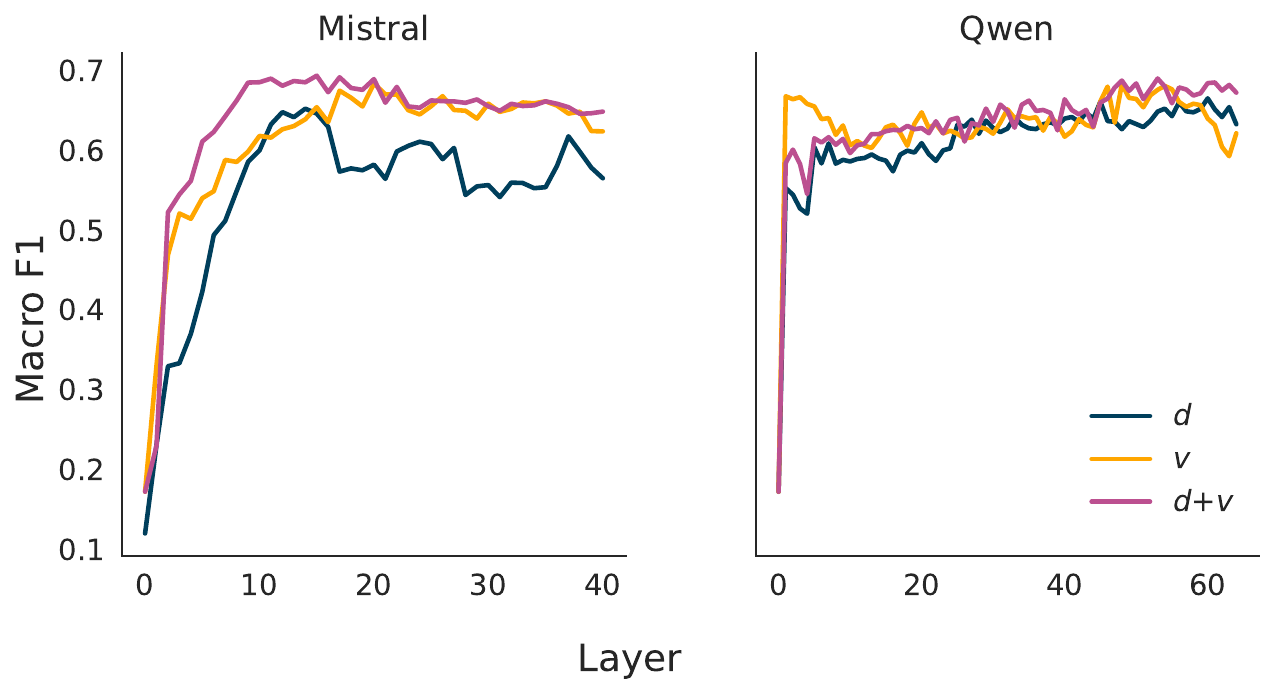}
    \caption{Layer-wise probe performance for Mistral and Qwen across digit (d), visual (v), and multimodal (d+v) inputs, with visual and multimodal representations yielding stronger separability.}\vspace{-0em}
    \label{fig:MistralQwenProbe}
\end{figure}

\vspace{2mm} \noindent \textbf{Layer Analysis.}
\emph{Information is encoded in early layers.}
In both plots of Figure~\ref{fig:MistralQwenProbe}, we see a quick improvement. Qwen encodes useful information in layers 1 and 2. Mistral is more gradual but encodes around layer 12 for $d$ and $d+v$ or layer 20 for $v$.
On the Llama side (Figure~\ref{fig:RandomLlamaProbe}), we observe Llama also peaking early, roughly at layer 5 on $d+v$ and at layer 10 on $d$ alone. We observe that the Random baselines in the same figure immediately return to their mean level at layer 1. 

\emph{Presence of $v$ boosts over random baseline.}
In Figure~\ref{fig:RandomLlamaProbe} we see in the $d$ modality that llama and random are intermixed. This compares with the clean separation of $d+v$, the modality in which, starting around layer 5, llama is consistently stronger across layers than the random baseline.

\vspace{2mm} \noindent \textbf{Ablations.}
Prompting is commonly augmented with in-context examples and chain-of-thought reasoning. We therefore include an ablation to assess the impact of these components. Table~\ref{tab:PromptingStrats} reports F1 scores averaged across modalities for different prompting strategies.

\emph{Chain-of-thought does not have much effect on representation quality.}
For Llama and Qwen, average F1 decreases when using CoT, from .609 to .598 and from .667 to .654, respectively. Mistral shows a small improvement, increasing from .638 to .648. Overall, these results indicate that CoT prompting does not reliably enhance the quality of the extracted representations. Similar trends are observed when using two-shot probing, which we report in the appendix.

\emph{Similarly, CoT does not consistently improve prompting.}
Llama benefits from CoT, improving from .174 with direct prompting to .220 with CoT. However, Mistral performs worse with CoT (.296) than with direct zero-shot prompting (.340). Qwen shows no meaningful difference, with CoT (.289) nearly matching direct prompting (.290). These results suggest that CoT benefits are model-dependent and not universal.

\emph{In-context examples do not improve prompting performance.}
Across models, in-context examples either degrade performance or provide negligible gains. For Llama, average F1 decreases from .174 in the zero-shot setting to .163 with two-shot prompting. For Qwen, performance drops from .290 to .280. Mistral shows a small decrease from .340 to .321. These results align with prior findings that few-shot prompting is often ineffective for classification-style tasks.

Taken together, these results indicate that zero-shot direct prompting provides a reasonable and representative evaluation setting. Neither few-shot examples nor in-context learning consistently improve representation quality or downstream prompting performance in this task.    

\emph{Results extend to text-only LLMs.}
To evaluate whether our findings depend on multimodal architectures, we repeat our experiments using text-only LLMs. Specifically, we evaluate Gemma and GPT-oss using the $d$-only modality. Table~\ref{LLMAblation} reports the corresponding probe-based and prompt-based results.

Across both models, probing substantially outperforms prompting. Gemma achieves an average F1 of .596 with probing, compared to .206 with prompting. GPT-oss shows a similar pattern, achieving .466 with probing and .202 with prompting. This gap is consistent across all datasets.

These results indicate that the advantages of probing over prompting are not specific to visual inputs or multimodal architectures. Instead, they reflect a more general limitation of prompting as an interface for extracting discriminative representations in time-series classification.

\label{prompt-stability}
\definecolor{passorange}{rgb}{1.0,0.94,0.90}
\definecolor{passblue}{rgb}{0.92,0.95,1.0}

\begin{table*}[t]
\centering
\footnotesize
\renewcommand{\arraystretch}{1.08}
\begin{tabular}{@{}ll
>{\columncolor{passorange}}c
>{\columncolor{passorange}}c
>{\columncolor{passorange}}c
>{\columncolor{passorange}}c
>{\columncolor{passorange}}c
>{\columncolor{passblue}}c
>{\columncolor{passblue}}c
>{\columncolor{passblue}}c
@{}}
\toprule
\textbf{dataset} & \textbf{modality} &
\textbf{min} & \textbf{max} & \textbf{mean} & \textbf{median} & \textbf{$\Delta$} &
\textbf{P@1} & \textbf{P@20}  &\textbf{$\Delta$P@K}\\
\midrule

\multirow{3}{*}{CTU}
 & $d$   & .410 & .509 & .463 & .470 & .099 & .506 & .568  &.098
\\
 & $d+v$ & .335 & .444 & .380 & .389 & .109 & .471 & .548  &.000
\\
 & $v$   & .357 & .434 & .394 & .395 & .077 & .457 & .556  &.052
\\
\midrule

\multirow{3}{*}{EMG}
 & $d$   & .167 & .395 & .202 & .167 & .228 & .343 & .400  &.057
\\
 & $d+v$ & .140 & .376 & .197 & .167 & .236 & .333 & .333  &.000
\\
 & $v$   & .167 & .167 & .167 & .167 & .000 & .333 & .333  &.000
\\
\midrule

\multirow{3}{*}{HAD}
 & $d$   & .013 & .048 & .031 & .031 & .035 & .043 & .100  &.062
\\
 & $d+v$ & .016 & .025 & .020 & .021 & .009 & .059 & .074  &.099
\\
 & $v$   & .037 & .056 & .049 & .050 & .019 & .163 & .198  &.077
\\
\midrule

\multirow{3}{*}{HAR}
 & $d$   & .122 & .151 & .138 & .138 & .029 & .175 & .278  &.103
\\
 & $d+v$ & .088 & .145 & .130 & .137 & .057 & .274 & .314  &.034
\\
 & $v$   & .080 & .157 & .143 & .152 & .077 & .310 & .344  &.040
\\
\midrule

\multirow{3}{*}{TEE}
 & $d$   & .000 & .081 & .027 & .016 & .081 & .140 & .238  &.057
\\
 & $d+v$ & .052 & .188 & .102 & .097 & .136 & .400 & .452  &.035
\\
 & $v$   & .038 & .167 & .110 & .102 & .129 & .357 & .357  &.015
\\
\midrule
\textbf{Avg ($d$)} & $d$
 & .142& .237& .172& .164
& \textbf{.094}& .242
& .317&\textbf{.075}\\

\textbf{Avg ($v$)} & $v$
 & .136& .196& .173& .173
& \textbf{.060}& .324
& .358&\textbf{.034}\\

\textbf{Avg ($d{+}v$)} & $d+v$
 & .126& .236& .166& .162
& \textbf{.109}& .308
& .344&\textbf{.037}\\

\midrule
\textbf{Total Avg} & --
 & .135& .223& .170& .166& \textbf{.088}& .291& .340&\textbf{.048}\\
\bottomrule\end{tabular}
\caption{Prompt stability results across datasets and modalities. Orange columns reflect sensitivity to prompt wording, and blue columns reflect variability from repeated sampling. Larger $\Delta$ (max - min F1) and $\Delta$P@K indicate less stable prompt-based evaluation.} \vspace{-0em}

\label{tab:variant_passk_combined}
\end{table*}

\begin{figure}[t]
    \centering
    \includegraphics[width=.9\linewidth]{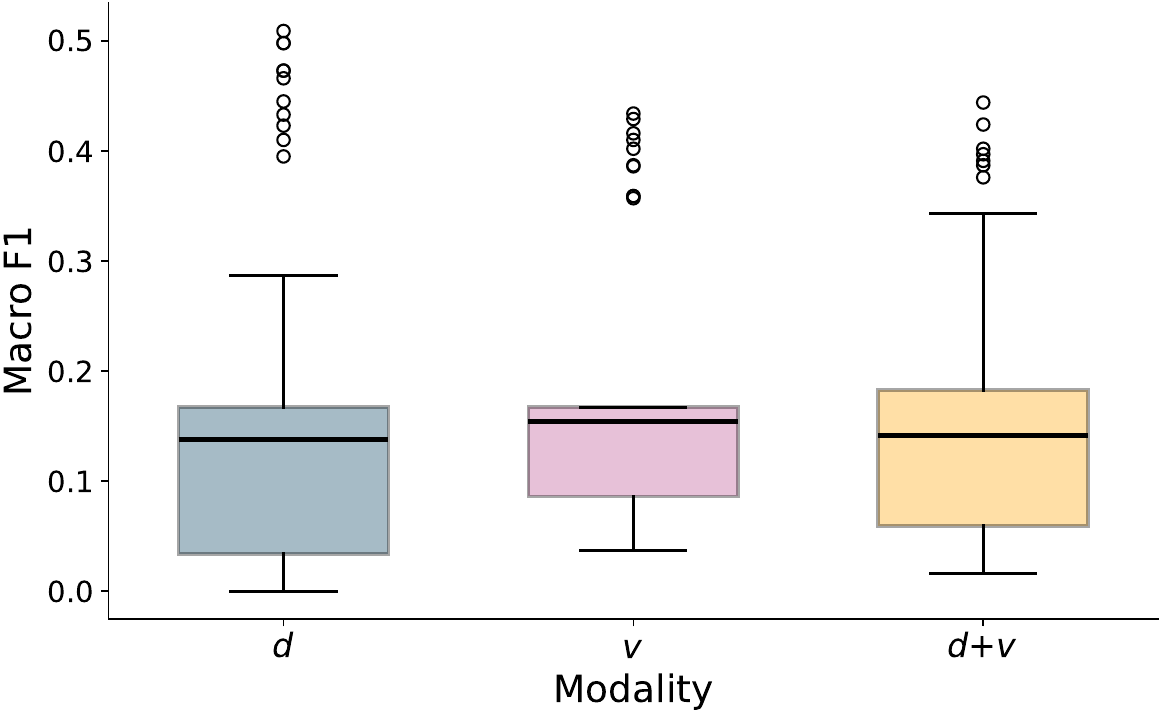}
    \caption{Prompt-based macro F1 distributions input modalities (d, v, d+v), showing a bit more stable performance for visual-only prompting.}\vspace{-1em}  
    \label{fig:BoxWhisp}
\end{figure}

\vspace{2mm} \noindent \textbf{Prompt Stability.}
We hypothesize that performance in LLM-based time series classification is highly sensitive to prompt engineering choices. We evaluate this hypothesis in Table~\ref{tab:variant_passk_combined}, Figure~\ref{fig:BoxWhisp}, which reports the results from two experiments.

First (orange-shaded columns), we assess \textit{sensitivity to prompt wording} by using \texttt{gpt-4.1-mini} to generate ten meaning-preserving variations of the original prompt. A larger \textbf{$\Delta$} (max$-$min macro F1) indicates greater variability in performance across prompt variants, and thus higher sensitivity to prompt phrasing. Across datasets and modalities, we observe an average $\Delta$ of .088 between the best- and worst-performing prompt variants, indicating substantial sensitivity even under minor prompt modifications. Examples of the prompt-variants can be found in Figures~\ref{fig:prompt_var_ex1}, \ref{fig:prompt_var_ex2}, \& \ref{fig:prompt_var_ex3} and the prompts used to generate them in Figures~\ref{fig:prompt_variation_system_templates} \& \ref{fig:prompt_variation_gq_templates}.

Second (blue-shaded columns), we evaluate \textit{output consistency} by prompting the model twenty times with the same prompt. The resulting P@K scores (more information found in Appendix~\ref{sec:appendix:BDMM:metrics}) measure the probability that at least one of $K$ sampled completions is correct. The corresponding $\Delta$P@K (P@20$-$P@1) quantifies the benefit of repeated sampling. On average, allowing twenty attempts increases accuracy by .048.

Notably, \textit{visual-only prompting ($v$) is the most stable modality under both criteria}. It exhibits the lowest average prompt-variant sensitivity ($\Delta = .060$) and the smallest sampling gain ($\Delta$P@K $= .034$), suggesting more consistent behavior both across prompt wording and repeated generations.

Together, these results support our claim that prompt-based evaluation in time series classification is highly sensitive to prompt engineering choices, and that reported performance may reflect prompt design and sampling strategy as much as underlying model capability.


\section{Conclusion}

This research provides a rigorous reassessment of LLMs' ability to represent and classify functional time-series data. By contrasting direct prompting with linear probing across diverse architectures and modalities, we identify a persistent gap in representation performance. Our evidence suggests that, although current prompting interfaces fail to capture temporal dynamics, the internal hidden states of these models contain highly discriminative, class-relevant information.
The results indicate that the perceived limitations of LLMs in time series tasks are primarily an artifact of the decoding interface rather than a fundamental deficiency in representational capacity. The early emergence of class discriminative signals, often within the first five transformer layers, suggests that these models function as effective general-purpose sequence encoders before higher-level reasoning stages.  Furthermore, the competitive performance of simple linear probes against specialized, state-of-the-art time series baselines underscores the quality of pretrained LLMs as robust feature extractors for non-linguistic modalities.

\section*{Limitations}
This paper is primarily diagnostic. We focus on measuring and explaining the gap between prompt-level performance and representational capacity, rather than proposing a new prompting method to close this gap. Developing more robust prompting strategies is an important direction for future work. While we take care to address standard probing concerns and follow established best practices (see Appendix~\ref{sec:appendix:probing}), probing remains an imperfect but informative diagnostic.

In particular, strong probe performance should not be interpreted as evidence of high-level temporal reasoning or semantic understanding. Given the nature of the evaluated datasets, it is plausible that probes primarily exploit relatively low-level temporal cues such as amplitude ranges, monotonic trends, periodicity, or short-range autocorrelation. Importantly, these cues are precisely the signals required for accurate classification in many standard time-series benchmarks. The fact that linear probes can reliably extract such task-relevant information, while prompt-based generation cannot, highlights a limitation of prompting as an evaluation interface rather than a deficiency in the underlying representations.

Our random-weight control is limited. We include a single random-weight model to rule out trivial effects from supervised probe training, but extending this analysis to additional random initializations and model families would further strengthen conclusions about architectural separability. Importantly, however, the qualitative gap between probing and prompting is large, consistent across datasets, modalities, and model families, and substantially exceeds the variance introduced by probe training itself, making it unlikely that our core findings are driven by idiosyncrasies of a particular random initialization.

Finally, our evaluation is necessarily scoped. We study six time-series datasets, five language and vision–language models, and a broad set of state-of-the-art and heuristic baselines. In fact, the consistency of the observed trends across these diverse settings suggests that the central result—that prompt-based evaluation systematically underestimates representational capacity—is not sensitive to any single dataset or model choice. Exploring additional datasets, representations, and model scales remains an important direction for future work.

\bibliography{custom}

\appendix

\section{Baselines, Datasets, Models \& Metrics}
\label{sec:appendix:additional_results}

\begin{table*}[t]
\centering

\begin{tabular}{@{}llllllllll@{}}
\toprule
\textbf{Method}               & \textbf{Model}   & \textbf{Shots} & \textbf{Style}  & \textbf{ctu}   & \textbf{emg}   & \textbf{har}   & \textbf{rwc}   & \textbf{tee}   & \textbf{AVG}\\
\midrule
\multirow{6}{*}{Probe} & \multirow{2}{*}{Llama}   & 0     & CoT    & .651 & .789 & .629 & .566 & .351 & .598       \\
                     &         &   0    & Direct & .650 & .784 & .645 & .608 & .355 & .609       \\ \cmidrule{2-10}
                     & \multirow{2}{*}{Mistral} & 0     & CoT    & .688 & .899 & .653 & .581 & .418 & .648       \\
                     &         &   0    & Direct & .674 & .871 & .655 & .585 & .408 & .638       \\ \cmidrule{2-10}
                     & \multirow{2}{*}{Qwen}    & 0     & CoT    & .697 & .920 & .677 & .595 & .442 & .654       \\
                     &         &   0    & Direct & .702 & .938 & .704 & .592 & .473 & .667       \\
\midrule
\multirow{9}{*}{Prompt}    & \multirow{3}{*}{Llama}   & 0     & CoT    & .434 & .233 & .126 & .231 & .077 & \textbf{.220}\\
                     &         &     0  & Direct & .333 & .185 & .083 & .231 & .036 & .174       \\
                     &         & 2     & Direct & .333 & .173 & .041 & .231 & .036 & .163       \\ \cmidrule{2-10}
                     & \multirow{3}{*}{Mistral} & 0      & CoT    & .544 & .324 & .168 & .339 & .105 & .296       \\
                     &         &    0   & Direct & .420 & .393 & .199 & .398 & .288 & \textbf{.340}\\
                     &         & 2     & Direct & .423 & .445 & .230 & .320 & .188 & .321       \\ \cmidrule{2-10}
                     & \multirow{3}{*}{Qwen}    & 0     & CoT    & .498 & .127 & .225 & .401 & .195 & .289       \\
                     &         &    0   & Direct & .359 & .393 & .127 & .288 & .283 & \textbf{.290}\\
                     &         & 2     & Direct & .472 & .360 & .228 & .234 & .107 & .280 \\
\bottomrule
\end{tabular}
\caption{F1 results between prompting and probing. We explore whether Chain-of-Thought (CoT) prompting and few-shot examples improve prompting performance, as well as whether CoT improves the representations during probing. \textit{Shots} represent the number of in-line examples per class in the dataset. For example \textit{har} has 6 possible classes, there are 2 examples per class totaling to 12 examples overall.  These results come from a smaller sample using 500 training rows and 100 test rows.}
\label{tab:PromptingStrats}
\end{table*}

\begin{table*}[t]
\centering 

\begin{tabular}{@{}lllllllll@{}}
\toprule
\textbf{Model}   & \textbf{Method}               & \textbf{ctu}   & \textbf{emg}   & \textbf{had} & \textbf{har}   & \textbf{rwc}   & \textbf{tee}   & \textbf{AVG}\\
\midrule
Gemma   & \multirow{2}{*}{Probe}                & .680& 1.00&     .145& .739& .556& .458& .596
\\
GPT-oss &                       & .579& .792&     .112& .498& .518& .295& .466
\\
\cmidrule{2-9}

Gemma   & \multirow{2}{*}{Prompt}                & .286& .314&     .023& .075& .346& .100& .206
\\
GPT-oss &                       & .282& .358&     .031& .142& .272& .132& .202\\
\bottomrule

\end{tabular}
\caption{F1 scores for probe-based and prompt-based inference on text-only large language models.}
\label{LLMAblation}
\end{table*}

\subsection{Baselines}
\label{sec:appendix:baselines}

\textbf{Attend} follows the dedicated time-series encoder family. It uses self-attention to model cross-channel and temporal dependencies in multivariate sensor data and learns discriminative representations directly from raw time series, without adapting language-model token spaces~\cite{Abedin2021Attend}. \textbf{InstructTime} belongs to the time-series quantization and instruction-based prompting family. It discretizes time series into vector-quantized tokens, aligns them with domain-specific textual instructions, and uses generative instruction tuning to produce label text via next-token prediction~\cite{Cheng2025InstrucTime}. \textbf{OneFitsAll} follows the encoder-alignment family. It reuses a frozen pretrained transformer from language or vision domains and learns lightweight input projections and normalization layers to map time-series patches into the model’s embedding space, enabling a single backbone across tasks~\cite{Zhou2023OneFitsAll}. \textbf{TS2Vec} represents dedicated time-series representation learning. It learns multi-scale contextual embeddings through hierarchical contrastive learning over augmented temporal views, which are aggregated and used by simple downstream classifiers without language models~\cite{yue2022ts2vecuniversalrepresentationtime}. \textbf{Moment} belongs to the TS-Encoder family. It is a high-capacity transformer encoder pre-trained on large-scale, multi-domain time series using masked time-series modeling, and serves as a general-purpose representation learner that can be adapted to classification, forecasting, anomaly detection, and imputation via lightweight task-specific heads or linear probing~\citep{goswami2024moment}.

\subsection{Models}
To assess the generality of our findings, we evaluate a diverse set of pretrained models spanning different architectures and scales.\footnote{All models are open source, as access to internal hidden states is required for probing.} Motivated by prior work showing that visualizations provide a strong representation for time series classification~\cite{liu-etal-2025-picture}, we primarily focus on vLLMs.
Specifically, we evaluate \texttt{Llama-3.2-11B-Vision-Instruct}~\cite{grattafiori2024llama3herdmodels},
\texttt{Qwen/Qwen2.5-VL-32B-Instruct}~\cite{bai2025qwen3vltechnicalreport},
and \texttt{Mistral-Small-3.1-24B-Instruct-2503}~\cite{MistralCard}. To verify that our conclusions extend beyond multimodal architectures, we also evaluate two text-only LLMs:
\texttt{GPT-oss}~\cite{openai2025gptoss120bgptoss20bmodel} and
\texttt{google/gemma-2-9b-it}~\cite{gemmateam2024gemma2improvingopen}.

\subsection{Datasets}
\label{sec:appendix:BDMM:A3Datasets}
Appendix~\ref{sec:appendix:stats} provides direct links to all source datasets used in our experiments, while Table~\ref{tab:datasets} reports descriptive statistics. 
We selected datasets to span a wide range of temporal scales, sequence lengths, variable dimensionality, and label complexity. This diversity allows us to evaluate whether LLM-based prompting generalizes across fundamentally different forms of temporal structure. Short descriptions of each dataset can be found below.

\textbf{CTU} records 24-hour household electricity usage patterns and classifies computers as desktops or laptops, capturing differences in device power-consumption cycles~\cite{UCRArchive}. \textbf{EMG} consists of muscle response recordings under electrical stimulation, categorized into healthy, myopathy, and neuropathy classes~\cite{goldberger2000physiobank}. \textbf{HAR} contains multivariate accelerometer recordings of human activities such as walking, sitting, and stair climbing, and exhibits substantial inter-subject variability~\cite{Anguita2013APD}. \textbf{HAD} similarly contains multivariate human activity data; we use both the tri-axis accelerometer data and tri-axis gyroscopic data~\citep{Zhang2012USC}. \textbf{RWC} comprises underwater acoustic recordings used to detect North Atlantic right whale calls in the presence of ocean noise~\cite{cox2006understanding}. \textbf{TEE} originates from the FORTE satellite and classifies transient electromagnetic events (e.g., lightning strokes) based on waveform signatures~\cite{eads2002genetic}.

\subsection{Metrics}
\label{sec:appendix:BDMM:metrics}
Let $\mathcal{D}=\{(x_i,y_i)\}_{i=1}^N$ denote a test set of $N$ labeled time series, where $y_i \in \{1,\dots,C\}$ and $\hat{y}_i$ is the predicted class. Let
$TP_c, FP_c, FN_c$ denote the counts of true positives, false positives, and false negatives for class $c$ under the one-vs-rest convention.

\vspace{2mm} \noindent \paragraph{Macro F1.}
For each class $c$, define precision and recall:
\[
P_c = \frac{TP_c}{TP_c + FP_c}, \qquad
R_c = \frac{TP_c}{TP_c + FN_c}.
\]
The classwise F1 is
\[
F1_c = \frac{2 P_c R_c}{P_c + R_c}.
\]
Macro F1 averages uniformly across classes:
\[
F1_{\text{macro}} = \frac{1}{C}\sum_{c=1}^{C} F1_c.
\]

\vspace{2mm} \noindent \paragraph{Pass@K.}
For each test example $i$, we obtain $n$ independent generations and mark each
generation as correct/incorrect. Let $c_i$ be the number of correct generations
among the $n$ samples for example $i$ (in our experiments, $n=20$). Following the
standard unbiased estimator for pass@K, we define:
\[
\widehat{\mathrm{Pass@}K}_i
= 1 - \frac{\binom{n - c_i}{K}}{\binom{n}{K}},
\]
with the conventions $\widehat{\mathrm{Pass@}K}_i=0$ if $c_i=0$, and
$\widehat{\mathrm{Pass@}K}_i=1$ if $n-c_i < K$.
We report the dataset-level score by averaging across examples:
\[
\widehat{\mathrm{Pass@}K}
= \frac{1}{N}\sum_{i=1}^{N} \widehat{\mathrm{Pass@}K}_i.
\]

\vspace{2mm} \noindent \paragraph{$\Delta$Pass@K.}
We quantify the benefit of repeated sampling by comparing multiple attempts
against a single attempt:
\[
\Delta\widehat{\mathrm{Pass@}K}
= \widehat{\mathrm{Pass@}K} - \widehat{\mathrm{Pass@}1}.
\]
In particular, we report $\Delta\widehat{\mathrm{Pass@}20}
= \widehat{\mathrm{Pass@}20} - \widehat{\mathrm{Pass@}1}$.

\section{Addressing Probing Concerns}
\label{sec:appendix:probing}

This section provides additional analyses and controls addressing known methodological concerns associated with probing classifiers. Following the recommendations of \citet{belinkov_probing_2022}, these experiments are intended to support the interpretation of probing results as diagnostics of representational capacity rather than causal explanations of model behavior.

\vspace{2mm} \noindent \textbf{Probe Complexity and Memorization.}
A central concern in probing-based analysis is that probe performance may reflect the capacity of the probe itself rather than information encoded in the underlying representations~\cite{belinkov_probing_2022}. To mitigate this risk, all probes used in this work are linear classifiers trained on frozen representations. We use logistic regression, one of the simplest and most interpretable classifiers, commonly employed in probing studies to assess linear extract-ability.

To further guard against probe memorization, we evaluate probe performance across layers and datasets and observe consistent trends rather than isolated improvements at specific layers. All probes are trained using five-fold cross-validation with grid search to select the regularization parameter $C$, and the maximum number of iterations is fixed to 1000. Together, these design choices reduce the likelihood that observed gains arise from overfitting by the probe and support their interpretation as reflecting systematic representational structure.

\vspace{2mm} \noindent \textbf{Baselines and Controls.}
Interpreting raw probing accuracy in isolation can be misleading, as even simple or randomized representations may support non-trivial classification performance. Prior work, therefore, emphasizes the importance of contextualizing probing results using appropriate baselines and reference models~\cite{conneau-etal-2018-cram}. 

As an absolute lower bound, we report random-assignment baselines, including majority, uniform, and prior-based predictors. These baselines characterize dataset properties such as the number of classes and class imbalance, and provide a reference point for interpreting probe performance.

Next, we follow the guidance of \citet{belinkov-etal-2017-neural} and \citet{conneau-etal-2018-cram}, \textbf{we train a classifier on an LLM with randomly initialized weights}. Specifically, we chose \texttt{Llama-3.2-11B-Vision-Instruct} as it can be directly compared to its fine-tuned version.

The randomly initialized model serves as a practical floor for probing-based approaches. It reflects how well a linear classifier can extract a signal from features without meaningful representations. This baseline, therefore, controls for the effect of supervised training applied after feature generation.

As an upper-bound reference, established baselines designed specifically to learn discriminative time-series representations, following prior probing work that contrasts frozen representations with task-optimized or state-of-the-art models~\cite{belinkov-etal-2017-neural, conneau-etal-2018-cram, liu-etal-2019-linguistic}.

\vspace{2mm} \noindent \textbf{Correlation Versus Causation.}
Early probing work often assumed that extractable information reflected model usage, but subsequent studies have demonstrated mismatches between probing performance and downstream task behavior~\cite{Vanmassenhove_Du_Way_2017, belinkov2019analysismethodsneurallanguage}. Accordingly, we avoid causal claims about model decision-making and interpret probing results strictly as evidence of representational capacity rather than usage.

\vspace{2mm} \noindent \textbf{Choice of Probed Properties.}
Another limitation of probing is its reliance on predefined properties, which constrains the scope of conclusions and may bias analysis toward expected features. In our setting, we probe directly for time-series class labels, treating linear separability with respect to the downstream classification task as the property of interest. This choice aligns with our goal of assessing whether LLM representations encode discriminative temporal structure relevant to classification, rather than targeting specific hand-engineered temporal features.

\section{Dataset Statistics}
\label{sec:appendix:stats}
Below in Table~\ref{tab:datasets} we report the descriptive statistics of each of the datasets used in this study. Links for each dataset can be found below.

\begin{table}[h]
\centering

\renewcommand{\arraystretch}{0.85}
\resizebox{\columnwidth}{!}{
\begin{tabular}{lccccccc}
\toprule
\textbf{Dataset} & \textbf{N-Channels} & \textbf{Series-Length} &
\textbf{N-Classes} & \textbf{Train} & \textbf{Test} \\
\midrule
cpu & 1 & 720  & 2 & 250      & 250      \\
emg & 1 & 1500 & 3 & 267      & 47       \\
had & 6 & 128  & 12 & 34779    & 7929      \\
har & 3 & 206  & 6 & 7{,}352  & 2{,}947  \\
rwc & 1 & 4000 & 2 & 25{,}499 & 4{,}501  \\
tee & 1 & 319  & 7 & 70       & 73       \\
\bottomrule
\end{tabular}}
\caption{Summary statistics for all time-series datasets used in our classification experiments.}
\label{tab:datasets}
\end{table}

\vspace{2mm} \noindent \textbf{Dataset Links}
\begin{itemize}
    \setlength\itemsep{0pt}
    \setlength\parsep{0pt}
    \item \textbf{CTU}: \url{https://www.timeseriesclassification.com/description.php?Dataset=Computers}
    \item \textbf{HAD}: \url{http://sipi.usc.edu/HAD/}
    \item \textbf{EMG}: \url{https://physionet.org/content/emgdb/1.0.0/}
    \item \textbf{HAR}: \url{https://archive.ics.uci.edu/dataset/240/human+activity+recognition+using+smartphones}
    \item \textbf{RWC}: \url{https://www.kaggle.com/competitions/whale-detection-challenge/data}
    \item \textbf{TEE}: \url{https://www.timeseriesclassification.com/description.php?Dataset=Lightning7}

\end{itemize}

\begin{figure}[t]
    \centering
    \includegraphics[width=\linewidth]{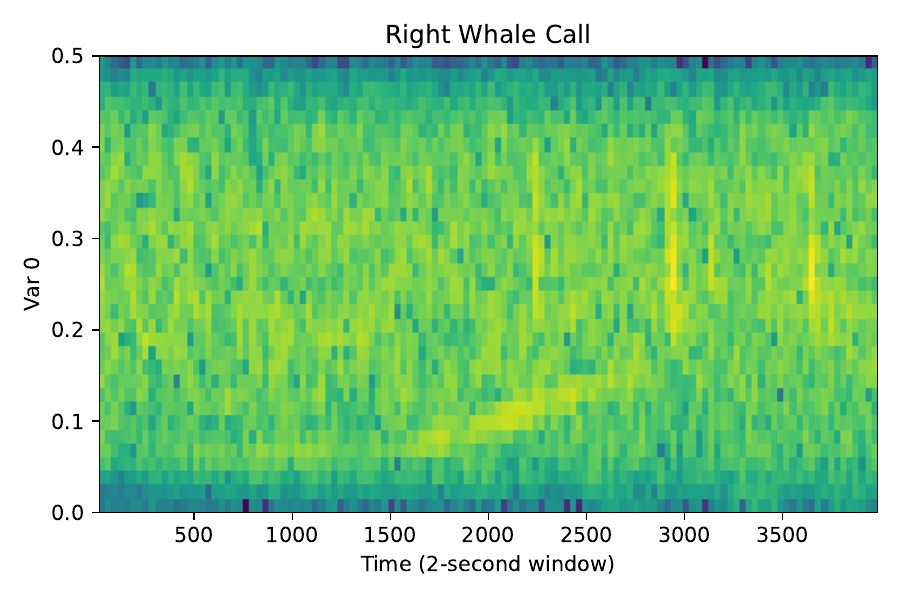}
    \caption{For the RWC dataset only, we represent the data as a spectrogram.}
    \label{fig:rwcplot}
\end{figure}

\section{Prompting Examples}
\label{sec:appendix:prompt_examples}

This section illustrates how time series data are visualized in the visual domain and how these representations are incorporated into the prompting structures used for evaluation. We show both line-plot and spectrogram-based visualizations, text-only ($d$) prompts for inference, prompts used to generate prompt-variants, and examples of prompt-variants themselves.

\subsection{Example Plots}
\label{sec:appendix:prompt_examples:ex_plots}

\begin{figure}[t]
    \centering
    \includegraphics[width=\linewidth]{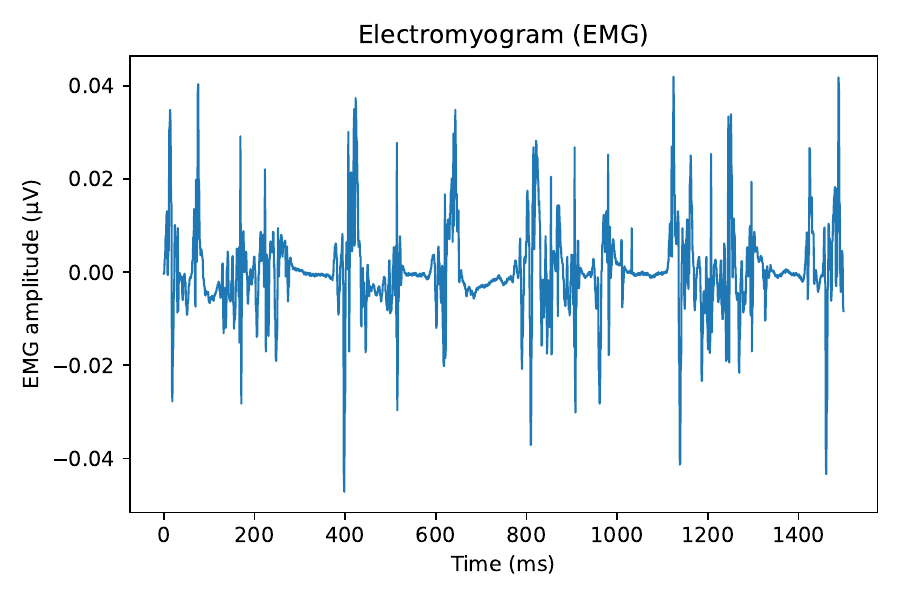}
    \caption{Example of a univariate time series line plot.}
    \label{fig:emgplot}
\end{figure}

\begin{figure}[t]
    \centering
    \includegraphics[width=\linewidth]{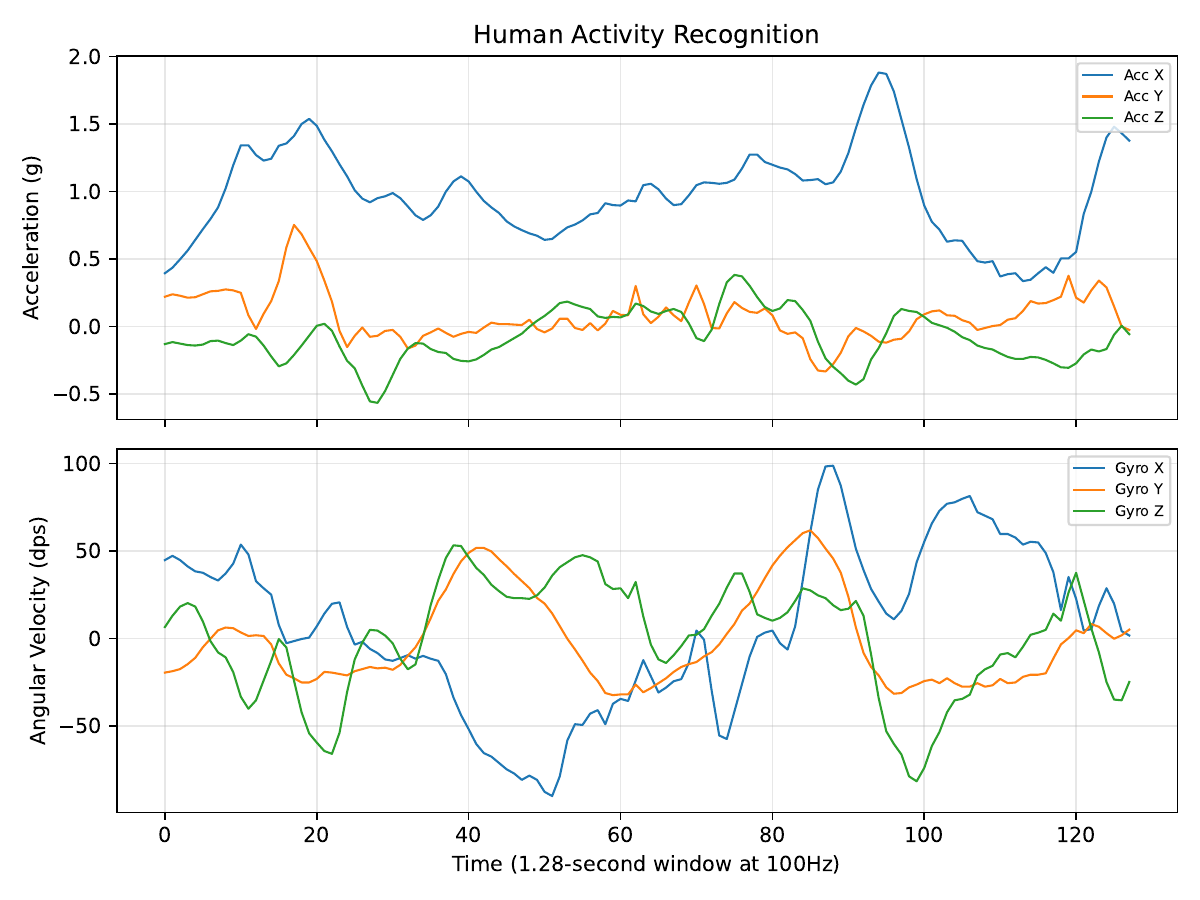}
    \caption{Example of a multivariate time series line plot.}
    \label{fig:hadplot}
\end{figure}

Figures~\ref{fig:rwcplot}, \ref{fig:emgplot}, \& \ref{fig:hadplot} demonstrate how time series information we represent time series in the visual domain. For the RWC dataset (Figure~\ref{fig:rwcplot}), following the recommendations of \citet{liu-etal-2025-picture}, the time series is rendered as a spectrogram. All other univariate datasets are visualized as line plots (see Figure~\ref{fig:emgplot} for an example). For multivariate datasets, each variable is plotted as a separate line with color-coded legends, expanding to multiple subplots when necessary (Figure\ref{fig:hadplot}). Axis labels, titles, and legends are specified by the user prior to inference.

\subsection{Prompting}
Below we show the prompting structure for inference as well as the prompt used to make the variants.

\paragraph{Prompts for Inference.}
Figure~\ref{fig:ctu_prompt_example} shows the instantiated prompt template used to classify the CTU dataset in the text-only ($d$) modality. In the visual ($v$) modality, the numerical time series is replaced with an appropriate visualization (see Section~\ref{sec:appendix:prompt_examples:ex_plots}). In the combined ($d{+}v$) modality, both the numerical sequence and the corresponding visualization are provided. The \textit{Additional information that may help} is specified \emph{a priori} and is adapted directly from \citet{liu-etal-2025-picture}, who demonstrated the effectiveness of this prompting strategy.

\paragraph{Prompts for Prompt-Variant Generation.}
Figures~\ref{fig:prompt_variation_system_templates} and \ref{fig:prompt_variation_gq_templates} illustrate the prompts used to generate diverse, meaning-preserving variants. Figure~\ref{fig:prompt_variation_system_templates} is used to generate variations of the system prompt, while Figure~\ref{fig:prompt_variation_gq_templates} is used to generate variations of the user message. For both system prompt and question generation, we issue two batch requests with batch size $B{=}5$, yielding a total of ten unique variants.

\paragraph{Examples of Prompt Variants.}
Figures~\ref{fig:prompt_var_ex1}, \ref{fig:prompt_var_ex2}, \& \ref{fig:prompt_var_ex3} present three prompt variants generated for the HAR dataset. Each variant exhibits distinct characteristics. For example, Variant~2 (Figure~\ref{fig:prompt_var_ex2}) is substantially more verbose than the others, whereas Variant~1 (Figure~\ref{fig:prompt_var_ex1}) is the most concise and omits expanded definitions for acronyms (e.g., using \texttt{RMS} without explicitly stating ``Root Mean Square,'' as done in the other examples).

\begin{figure*}[t]
\footnotesize
\centering

\begin{tcolorbox}[
  colback=gray!5!white,
  colframe=gray!75!black,
  title=System Prompt,
  width=\textwidth
]
\textbf{Play as a computer energy consumption analysis expert:} determine whether this computer is a desktop or a laptop based on the 24-hour power consumption time series.

\vspace{0.5em}
You will be given a multiple choice question and a time series. Your job is to use the time series to answer the question. \textbf{Do not explain your reasoning.}
Use exactly this format: \texttt{The answer is [X] CLASS\_NAME}.
Example: \texttt{The answer is [D] CABBAGE}.

\vspace{0.75em}
\noindent\textbf{Additional information that may help:}
\begin{itemize}\setlength\itemsep{0.2em}
  \item \textbf{Total daily consumption:} laptops typically consume less power than desktops.
  \item \textbf{Patterns over time:} laptops may show more uniform usage (sleep/hibernate); desktops may show sharper on/off contrasts.
  \item \textbf{Spikes:} desktops may exhibit higher spikes during active use.
  \item \textbf{Minimum/baseline:} laptops often have lower baselines; desktops may drop near zero if off or stay higher with peripherals.
  \item \textbf{Charging cycles:} periodic rises/falls consistent with battery charging suggest a laptop.
  \item \textbf{Variability:} laptops often have lower variance; desktops may fluctuate more across modes.
  \item \textbf{Day/night behavior:} strong overnight reductions may indicate sleep behavior typical of laptops.
\end{itemize}
\end{tcolorbox}

\vspace{0.9em}

\begin{tcolorbox}[
  colback=gray!5!white,
  colframe=gray!75!black,
  title=User Prompt,
  width=\textwidth
]
\noindent\textbf{Dataset:} \texttt{CTU}

\vspace{0.5em}
\noindent\textbf{Question:} determine whether this computer is a desktop or a laptop based on the 24-hour power consumption time series.

\vspace{0.5em}
\noindent\textbf{Options:}
\begin{itemize}\setlength\itemsep{0.15em}
  \item \texttt{[A] Desktop}
  \item \texttt{[B] Laptop}
\end{itemize}

\vspace{0.5em}
\noindent\textbf{Time series:} \\
\texttt{
0 2 0 4, 0 3 9 5, 0 2 0 4, ... , 1 7 0 9, 1 3 2 6, 2 2 8 3
}

\vspace{0.5em}
\noindent\textbf{Output format requirement:} \texttt{The answer is [X] CLASS\_NAME}
\end{tcolorbox}

\caption{System and user prompts for the CTU dataset. The time series is abbreviated for readability. The \textit{Additional information that may help}, \textit{question}, and \textit{options} change depending on the dataset.}
\label{fig:ctu_prompt_example}
\end{figure*}

\begin{figure*}[t]
\footnotesize
\centering

\begin{tcolorbox}[
  colback=gray!5!white,
  colframe=gray!75!black,
  title=Variation Generator: System Prompt Rewriter (System Prompt),
  width=\textwidth
]
You are a prompt rewriter.

Rewrite the provided SYSTEM PROMPT into N distinct SYSTEM PROMPT variants that
preserve meaning, constraints, and all factual task content, while changing
phrasing, structure, and formatting.

\vspace{0.4em}
\noindent\textbf{Hard constraints (must obey):}
\begin{itemize}\setlength\itemsep{0.15em}\setlength\parskip{0pt}\setlength\topsep{0pt}
  \item Do NOT add, remove, rename, or reorder any class names.
  \item Do NOT alter the semantic meaning of any class description.
  \item Preserve answer-choice letters (e.g., [A], [B]) exactly and in order.
  \item Preserve required output format strings exactly (punctuation/casing),
        e.g., \texttt{"The answer is [X] CLASS\_NAME"}.
  \item Do NOT introduce new task instructions (e.g., reasoning, confidence).
  \item Maintain an academically appropriate tone.
\end{itemize}

\vspace{0.2em}
\noindent\textbf{Diversity requirements:}
\begin{itemize}\setlength\itemsep{0.15em}\setlength\parskip{0pt}\setlength\topsep{0pt}
  \item Each variant must differ meaningfully in organization and wording.
  \item Use multiple presentation styles (headings, bullets, numbered steps, etc.).
  \item Avoid trivial paraphrases.
\end{itemize}

\vspace{0.2em}
\noindent\textbf{Output format (strict):} return valid JSON only:
\begin{verbatim}
{
  "variants": [
    {"id": 1, "system_prompt": "..."},
    {"id": 2, "system_prompt": "..."}
  ]
}
\end{verbatim}
No extra keys. No markdown. No commentary.
\end{tcolorbox}

\vspace{0.7em}

\begin{tcolorbox}[
  colback=gray!5!white,
  colframe=gray!75!black,
  title=Variation Generator: System Prompt Rewriter (User Message Template),
  width=\textwidth
]
\noindent\textbf{Per-batch request (batch size = $B$):}
\begin{verbatim}
Generate B rewritten variants of the following.

IMPORTANT:
- Do NOT change class names.
- Do NOT change answer letters or their order.
- Do NOT change the required answer format.
- Output must be valid JSON only.

<BASE_SYSTEM_PROMPT>
\end{verbatim}
\vspace{0.2em}
\noindent\textbf{Batching:} if total variants = $N$ and batch size = $B$, the script makes $N/B$ calls and then renumbers variant ids globally.
\end{tcolorbox}
\vspace{0.7em}
\caption{System and user prompts for system prompt variation generation. \texttt{<BASE\_SYSTEM\_PROMPT>} is replaced by the system prompt shown in Figure~\ref{fig:ctu_prompt_example} or by the dataset-specific base system prompt used for evaluation. Two batch requests with batch size $B{=}5$ are used, producing ten total system prompt variants.}
\label{fig:prompt_variation_system_templates}
\end{figure*}

\begin{figure*}[t]
\footnotesize
\centering

\begin{tcolorbox}[
  colback=gray!5!white,
  colframe=gray!75!black,
  title=Variation Generator: General Question Rewriter (System Prompt),
  width=\textwidth
]
You are a question rewriter.

Your task is to rewrite a GENERAL QUESTION into N distinct variants that preserve
the task, label space, and decision criteria, while changing wording, structure,
and phrasing.

\vspace{0.4em}
\noindent\textbf{Hard constraints (must obey):}
\begin{itemize}\setlength\itemsep{0.15em}\setlength\parskip{0pt}\setlength\topsep{0pt}
  \item Do NOT change, rename, remove, or reorder answer choices.
  \item Preserve answer-choice letters exactly (e.g., [A], [B]) and their order.
  \item Do NOT introduce new labels, hints, or constraints.
  \item Do NOT add explanations, reasoning instructions, or output formatting rules.
  \item The rewritten question must ask the same classification decision.
\end{itemize}

\vspace{0.2em}
\noindent\textbf{Diversity requirements:}
\begin{itemize}\setlength\itemsep{0.15em}\setlength\parskip{0pt}\setlength\topsep{0pt}
  \item Each variant must differ meaningfully in wording and structure.
  \item Vary tone (instructional vs.\ role-based), sentence structure, and framing.
  \item Avoid trivial paraphrases.
\end{itemize}

\vspace{0.2em}
\noindent\textbf{Output format (strict):} return valid JSON only:
\begin{verbatim}
{
  "variants": [
    {"id": 1, "question": "..."},
    {"id": 2, "question": "..."}
  ]
}
\end{verbatim}
No extra keys. No markdown. No commentary.
\end{tcolorbox}

\vspace{0.7em}

\begin{tcolorbox}[
  colback=gray!5!white,
  colframe=gray!75!black,
  title=Variation Generator: General Question Rewriter (User Message Template),
  width=\textwidth
]
\noindent\textbf{Per-batch request (batch size = $B$):}
\begin{verbatim}
Generate B rewritten variants of the following.

IMPORTANT:
- Do NOT change the answer choice letters, names, or their order.
- Do NOT add output formatting rules.
- Do NOT add reasoning/explanation instructions.
- Output must be valid JSON only.

<BASE_GENERAL_QUESTION>
\end{verbatim}
\vspace{0.2em}
\noindent\textbf{Batching:} if total variants = $N$ and batch size = $B$, the script makes $N/B$ calls and then renumbers variant ids globally.
\end{tcolorbox}

\caption{System and user prompts for general question variation generation. \texttt{<BASE\_GENERAL\_QUESTION>} is replaced by the dataset-specific general question used for evaluation (e.g., \textit{determine whether this computer \ldots} from Figure~\ref{fig:ctu_prompt_example}). As in system prompt variation generation, we issue two batch requests with batch size $B{=}5$ to obtain ten total question variants.}
\label{fig:prompt_variation_gq_templates}
\end{figure*}


\begin{figure*}[t]
\centering
\footnotesize

\begin{tcolorbox}[
  colback=gray!5!white,
  colframe=gray!75!black,
  title=Example Prompt Variations (Feature list with enumerated structure),
  width=\textwidth,
  boxsep=3pt,
  left=4pt,
  right=4pt,
  top=4pt,
  bottom=4pt,
  breakable=false
]

\noindent\textbf{Variant 1.}

{\scriptsize
\begin{verbatim}
You are tasked with human activity recognition based on tri-axial accelerometer
data (x, y, z) collected over time. Presented with a multiple-choice question
and corresponding time series data, your responsibility is to select the correct
activity. Provide your answer strictly in the following format:
"The answer is [X] CLASS_NAME" (e.g., "The answer is [D] CABBAGE"), without any
explanations.

Key features to consider for accurate classification include:
1. Statistical: mean, standard deviation, variance, median, range, IQR
2. Time-domain: RMS, zero crossing rate, signal magnitude area, time between peaks
3. Frequency-domain: FFT, PSD, spectral entropy, peak frequency
4. Correlation-based: correlations between axes
5. Geometrical: angles between axes, magnitude of acceleration vector
6. Entropy-based: signal entropy

Utilize these feature types to interpret the accelerometer signals and determine
the human activity. Analyze the sequence of tri-axial accelerometer measurements
(x, y, z) collected over time and classify the observed human activity into one
of these categories:
[A] WALKING [B] WALKING_UPSTAIRS [C] WALKING_DOWNSTAIRS [D] SITTING [E] STANDING [F] LAYING
\end{verbatim}
}

\end{tcolorbox}

\caption{Example one of three meaning-preserving prompt variations for human activity recognition.}
\label{fig:prompt_var_ex1}
\end{figure*}


\begin{figure*}[t]
\centering
\footnotesize

\begin{tcolorbox}[
  colback=gray!5!white,
  colframe=gray!75!black,
  title=Example Prompt Variations (Sectioned guidance with descriptive bullets),
  width=\textwidth,
  boxsep=3pt,
  left=4pt,
  right=4pt,
  top=4pt,
  bottom=4pt,
  breakable=false
]

\noindent\textbf{Variant 2.}

{\scriptsize
\begin{verbatim}
As an expert in human activity recognition, your task is to identify the activity
from the tri-axial accelerometer data (x, y, z) recorded over time. You will
receive a multiple-choice question along with the time series data. Use the time
series to select the correct answer without providing any explanation. Respond
strictly in the following format: "The answer is [X] CLASS_NAME". For example:
"The answer is [D] CABBAGE".

Additional guidance:
When analyzing accelerometer data for activity recognition, consider extracting
the following features:

1. Statistical Features:
- Mean: Average acceleration per axis. - Standard Deviation: Variation in acceleration values.
- Variance: Spread of data points around the mean. - Median: Middle value splitting the data.
- Range: Difference between max and min values.
- Interquartile Range (IQR): Spread between 25th and 75th percentiles.
2. Time-domain Features:
- Root Mean Square (RMS): Magnitude of acceleration. - Zero Crossing Rate (ZCR): Frequency of signal sign changes.
- Signal Magnitude Area (SMA): Energy expenditure over time. - Time between Peaks: Duration between local maxima.
3. Frequency-domain Features:
- Fast Fourier Transform (FFT): Frequency components analysis.
- Power Spectral Density (PSD): Power distribution across frequencies.
- Spectral Entropy: Complexity of frequency distribution.
- Peak Frequency: Dominant frequency component.
4. Correlation-based Features:
- Correlation between Axes: Coordination of movement.
5. Geometrical Features:
- Angles between Axes: Body orientation.
- Magnitude of Acceleration Vector: Overall acceleration magnitude.
6. Entropy-based Features:
- Signal Entropy: Complexity of the acceleration signal.

Using the tri-axial accelerometer data (x, y, z) recorded over time, identify
which human activity is being performed. Choose from the following options:
[A] WALKING [B] WALKING_UPSTAIRS [C] WALKING_DOWNSTAIRS [D] SITTING [E] STANDING [F] LAYING
\end{verbatim}
}
\end{tcolorbox}

\caption{Example two of three meaning-preserving prompt variations for human activity recognition.}
\label{fig:prompt_var_ex2}
\end{figure*}


\begin{figure*}[t]
\centering
\footnotesize

\begin{tcolorbox}[
  colback=gray!5!white,
  colframe=gray!75!black,
  title=Example Prompt Variations (Compact rubric-style presentation),
  width=\textwidth,
  boxsep=3pt,
  left=4pt,
  right=4pt,
  top=4pt,
  bottom=4pt,
  breakable=false
]

\noindent\textbf{Variant 3.}

{\scriptsize
\begin{verbatim}
As a professional in human activity recognition, your responsibility is to
identify the activity from tri-axial accelerometer data (x, y, z) recorded over
time. You will be provided with a multiple-choice question and the time series
data. Use the data to select the correct answer without any explanation,
adhering strictly to the format: "The answer is [X] CLASS_NAME". Example:
"The answer is [D] CABBAGE".

Consider these feature types when analyzing the accelerometer signals:
1) Statistical Features: mean, standard deviation, variance, median, range,
   interquartile range.
2) Time-domain Features: root mean square (RMS), zero crossing rate (ZCR),
   signal magnitude area (SMA), time between peaks.
3) Frequency-domain Features: fast Fourier transform (FFT), power spectral
   density (PSD), spectral entropy, peak frequency.
4) Correlation-based Features: correlation between axes.
5) Geometrical Features: angles between axes, magnitude of acceleration vector.
6) Entropy-based Features: signal entropy.

Acting as a specialist in human activity recognition, use the tri-axial
accelerometer readings (x, y, z) collected over time to classify the performed
activity. Your options are:
[A] WALKING [B] WALKING_UPSTAIRS [C] WALKING_DOWNSTAIRS [D] SITTING [E] STANDING [F] LAYING
\end{verbatim}
}

\end{tcolorbox}

\caption{Example three of three meaning-preserving prompt variations for human activity recognition.}
\label{fig:prompt_var_ex3}
\end{figure*}

\twocolumn

\clearpage

\section{Hyperparameters}
wFor reproducibility, we provide the complete hyperparameter configurations for all baseline methods. These settings were either adopted from the original publications or carefully tuned to ensure fair comparison across all methods on our datasets.
\subsection{OneFitsAll}
All deep learning models are implemented in PyTorch using the Hugging Face Transformers library~\cite{wolf-etal-2020-transformers}. We employ a GPT-2-based architecture~\cite{radford2019language} where the pre-trained GPT-2 model serves as the backbone, with most parameters frozen, only the layer normalization and positional embedding layers remain trainable. The input time series are processed using a patching strategy, where patches are embedded into a high-dimensional space before being fed to the transformer layers. Training is performed using the RAdam optimizer~\cite{Liu2020OnTheVariance} with cross-entropy loss, and model performance is evaluated using 5-fold stratified cross-validation on the training set. Standardization normalization is applied to input features, and the model is evaluated on a separate held-out test set using accuracy as the key metric. All hyperparameters are summarized in Table~\ref{tab:OneFitsAll_hyperparameters}.

\begin{table}[t]
\centering
\resizebox{\columnwidth}{!}{
\begin{tabular}{lc}
\toprule
\textbf{Hyperparameter} & \textbf{Value} \\
\midrule
Model Architecture & GPT-2 (6 layers, pre-trained) \\
Model Dimension & 768 \\
Patch Size & 16 \\
Stride & 16 \\
Optimizer & RAdam \\
Learning Rate & 0.001 \\
Batch Size & 64 \\
Training Epochs & 50 \\
Dropout & 0.1 \\
Loss Function & Cross-Entropy \\
\bottomrule
\end{tabular}}
\caption{Hyperparameters and training configuration for OneFitsAll model}
\label{tab:OneFitsAll_hyperparameters}
\end{table}

\subsection{Attend}
We implement our experiments using PyTorch. Our network is trained end-to-end using the Adam optimizer~\cite{kingma2017adammethodstochasticoptimization} with a learning rate that decays periodically. The training data is partitioned into segments using a sliding window approach with 50\% overlap between adjacent windows. We employ mixup augmentation as a regularization technique and incorporate center loss to encourage intra-class compactness in the learned feature space. Dropout is applied at multiple stages of the network, including the RNN layers and the classifier, to prevent overfitting. Model performance is evaluated using 5-fold cross-validation to ensure robust and reliable results. All hyperparameters are summarized in Table~\ref{tab:Attend_hyperparameters}.

\begin{table}[t]
\centering
\begin{tabular}{lc}
\toprule
\textbf{Hyperparameter} & \textbf{Value} \\
\midrule
Epochs & 300 \\
Batch size & 256 \\
Learning rate & $10^{-3}$ \\
LR decay factor & 0.9 \\
LR decay step & 10 epochs \\
Sliding window ($W$) & 24 \\
Window overlap & 50\% \\
Mixup $\alpha$ & 0.8 \\
Center loss weight ($\beta$) & 0.3 \\
Dropout (features) & 0.5 \\
Dropout (RNN) & 0.25 \\
Dropout (classifier) & 0.5 \\
Hidden dimension & 128 \\
Filter size & 5 \\
Number of filters & 64 \\
\bottomrule
\end{tabular}
\caption{Training hyperparameters for Attend and Discriminate model}
\label{tab:Attend_hyperparameters}
\end{table}

\subsection{InstructTime}
We implement InstructTime using Python 3.12.7 and PyTorch 2.7.1 on an Nvidia H200 GPU. The framework consists of two stages: (1) a TStokenizer (VQ-VAE)~\cite{van2017neural} for time series discretization, and (2) an InstructTime model built upon the pre-trained GPT-2 language model~\cite{radford2019language} for classification. For the TStokenizer, we employ standard VQ-VAE training settings with mean pooling. For the InstructTime model, we use AdamW optimizer~\cite{loshchilov2017sgdrstochasticgradientdescent} with weight decay and employ a cosine annealing learning rate scheduler with warmup~\cite{loshchilov2017sgdrstochasticgradientdescent}. Following prior work on pre-trained language models, we use a smaller learning rate and batch size compared to training from scratch, as the PLM has been pre-trained with a large number of parameters. Mixed-precision training~\cite{micikevicius2018mixedprecisiontraining} is enabled using PyTorch's automatic mixed precision (AMP) to accelerate computation. The detailed hyperparameter settings for both stages are summarized in Table~\ref{tab:InstructTime_hyperparameters}.

\begin{table}[t]
\centering

\begin{tabular}{lc}
\toprule
\textbf{Hyperparameter} & \textbf{Value} \\
\midrule
\multicolumn{2}{c}{\textit{TStokenizer (VQ-VAE)}} \\
\midrule
Hidden dimension ($d_{\text{model}}$) & 64 \\
Codebook size ($n_{\text{embed}}$) & 256 \\
Wave length & 24 \\
Training epochs & 60 \\
Batch size (train/test) & 32 / 64 \\
Learning rate & $5 \times 10^{-4}$ \\
Learning rate decay & 0.99 \\
Weight decay & \(1e^{-5}\) \\
\midrule
\multicolumn{2}{c}{\textit{InstructTime}} \\
\midrule
Batch size & 16 \\
Learning rate & \(1e^{-5}\) \\
Weight decay & 0.01 \\
Warmup ratio & 0.05 \\
Training epochs & 300 \\
Max sequence length & 230 \\
Random seed & 2024 \\
\bottomrule
\end{tabular}
\caption{Hyperparameter settings for TStokenizer and InstructTime models}
\label{tab:InstructTime_hyperparameters}
\end{table}

\subsection{Moment Embedding Extraction}
We extract fixed-length embeddings using the pretrained \texttt{AutonLab/MOMENT-1-large} pipeline. Each time series is z-normalized using a per-channel \texttt{StandardScaler} fit on the training split and reused for the test split. To handle sequences longer than the model window, each series is segmented into non-overlapping windows of length $W{=}512$ with left-padding for the final partial window; an input mask marks padded positions. MOMENT outputs patch embeddings with patch length $8$, which we average over the valid (non-padded) patches to obtain a window embedding. Window embeddings are then aggregated into a single series embedding using a length-weighted average across windows. All configuration values are summarized in Table~\ref{tab:moment_hyperparameters}.
\begin{table}[t]
\centering
\small
\begin{tabular}{lc}
\toprule
\textbf{Hyperparameter} & \textbf{Value} \\
\midrule
Backbone model & \texttt{AutonLab/MOMENT-1-large} \\
Task name & classification \\
Series batch size & 64 \\
Window length ($W$) & 512 \\
Window stride & 512 \\
Patch length & 8 \\
\bottomrule
\end{tabular}
\caption{Hyperparameters for Moment  baseline.}
\label{tab:moment_hyperparameters}
\end{table}

\subsection{TS2Vec}
We implement TS2Vec using the authors’ official PyTorch implementation, and, with the exception of epochs, used a fixed set of hyperparameters across all datasets. We used 10 training epochs for larger datasets (HAD, RWC, HAR) and 30 epochs for our smaller datasets (CTU, EMG, TEE). Input time series are standardized prior to training, and representations are extracted from the final encoder layer for downstream evaluation. All hyperparameters are summarized in Table~\ref{tab:TS2Vec_hyperparameters}.

\begin{table}[t]
\centering
\resizebox{\columnwidth}{!}{
\begin{tabular}{lc}
\toprule
\textbf{Hyperparameter} & \textbf{Value} \\
\midrule
Representation Dimension & 320 \\
Hidden Channels & 64 \\
Residual Blocks & 10 \\
Kernel Size & 3 \\
Batch Size & 8 \\
Crop Length & 3000 \\
Normalization & Z-score (per channel) \\
Learning Rate & .0001 \\
Epochs & 10 or 30 \\
\bottomrule
\end{tabular}}
\caption{Hyperparameters and training configuration for TS2Vec.}
\label{tab:TS2Vec_hyperparameters}
\end{table}

\section{Use of AI Assistants}
We used AI to help clean up writing, but all thoughts and work are our own.

\end{document}